\documentclass[letterpaper, 10 pt, conference]{ieeeconf}  

\IEEEoverridecommandlockouts                              

\usepackage{graphicx}
\usepackage{hyperref}
\usepackage{amsmath}
\usepackage{amssymb}
\usepackage{booktabs}
\usepackage{multirow}
\usepackage{array}
\usepackage{balance}

\title{\LARGE \bf
EquiBim: Learning Symmetry-Equivariant Policy for Bimanual Manipulation
}

\author{Zhiyuan Zhang, Aditya Mohan, Seungho Han, Wan Shou, Dongyi Wang, Yu She$^{\dagger}$
\thanks{This work was partially supported by the United States Department of Agriculture (USDA) under Grant Nos. 2023-67021-39072 (Y.S.), 2023-67022-39074 (W.S.), 2023-67022-39075 (D.W.), and 2024-67021-42878 (Y.S.), and by the National Science Foundation (NSF) under Grant No. 2423068 (Y.S.).}
\thanks{$^{\dagger}$ Corresponding author.}
\thanks{Zhiyuan, Aditya, Seungho, and Yu are with Purdue University, West Lafayette, IN 47907,USA  
{\tt\footnotesize \{zhan5570, mohan81, han825, shey\}@purdue.edu}}
\thanks{Wan and Dongyi are with University of Arkansas, Fayetteville, AR 72701, USA  
{\tt\footnotesize \{wshou, dongyiw\}@uark.edu}}%
}

\begin{document}
\maketitle
\thispagestyle{empty}
\pagestyle{empty}

\begin{abstract}
Robotic imitation learning has achieved impressive success in learning complex manipulation behaviors from demonstrations.
However, many existing robot learning methods do not explicitly account for the physical symmetries of robotic systems, often resulting in asymmetric or inconsistent behaviors under symmetric observations.
This limitation is particularly pronounced in dual-arm manipulation, where bilateral symmetry is inherent to both the robot morphology and the structure of many tasks.
In this paper, we introduce EquiBim, a symmetry-equivariant policy learning framework for bimanual manipulation that enforces bilateral equivariance between observations and actions during training.
Our approach formulates physical symmetry as a group action on both observation and action spaces, and imposes an equivariance constraint on policy predictions under symmetric transformations.
The framework is model-agnostic and can be seamlessly integrated into a wide range of imitation learning pipelines with diverse observation modalities and action representations, including point cloud-based and image-based policies, as well as both end-effector-space and joint-space parameterizations.
We evaluate EquiBim on RoboTwin, a dual-arm robotic platform with symmetric kinematics, and validate it across diverse observation and action configurations in simulation.
We further validate the approach on a real-world dual-arm system.
Across both simulation and physical experiments, our method consistently improves performance and robustness under distribution shifts.
These results suggest that explicitly enforcing physical symmetry provides a simple yet effective inductive bias for bimanual robot learning.
For more details, please refer to the project website: \url{https://zhangzhiyuanzhang.github.io/equibim-website/}.
\end{abstract}

\section{INTRODUCTION}
Robotic imitation learning has become a powerful paradigm for acquiring complex manipulation skills directly from demonstrations, enabling robots to perform tasks that are difficult to model analytically~\cite{osa2018algorithmic, diffusion_policy, zhao2023learning}.
Recent advances have demonstrated strong performance across a wide range of manipulation scenarios, including dexterous manipulation~\cite{zhao2024aloha, an2025dexterous}, high-precision operations such as insertion and assembly~\cite{wu2025tacdiffusion, zhang2025vtla, luu2025manifeel}, and bimanual coordination~\cite{heng2025vitacformer, jiang2025dexmimicgen}.
These methods have benefited from increasingly expressive policy representations and richer sensory inputs, allowing robots to reason about complex scene geometry and multi-object interactions.

\begin{figure}[t]
    \centering
    \includegraphics[width=1.0\linewidth]{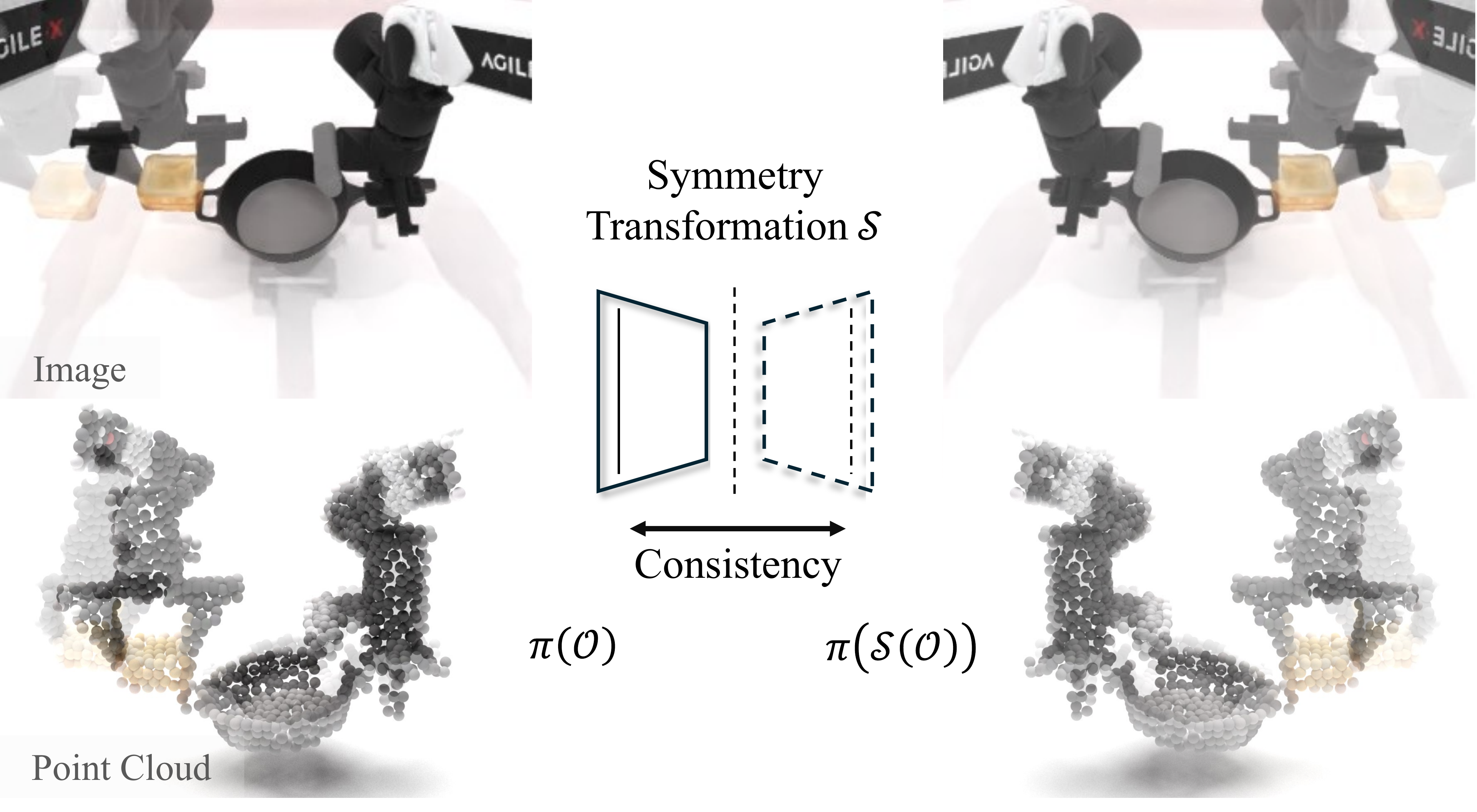}
    \caption{
    \textbf{Symmetry-equivariant policy learning for bimanual manipulation.}
    A symmetry transformation $\mathcal{S}$ defines a left--right exchange of the scene in the image coordinate frame, generating a symmetrically equivalent task instance.
    Given an observation $\mathcal{O}$ and its transformed counterpart $\mathcal{S}(\mathcal{O})$, the shared policy $\pi$ is trained to produce equivariant predictions under the same transformation, i.e.,
    $\pi(\mathcal{S}(\mathcal{O})) \approx \mathcal{S}(\pi(\mathcal{O}))$.
    The symmetry constraint is imposed at the action level and does not require architectural modifications.
    }
    \label{fig:abstract}
\end{figure}

Despite substantial progress in robotic imitation learning, most existing approaches do not explicitly enforce the physical symmetries that naturally arise in robotic systems and manipulation tasks~\cite{zhang2025canonical}.
This limitation is particularly evident in dual-arm manipulation, where bilateral symmetry is intrinsic to both robot morphology and actuation.
Many bimanual tasks, such as assembly, handover, and coordinated object transport, admit symmetric configurations and solution strategies, as illustrated in Fig.~\ref{fig:abstract}.
Nevertheless, policies trained with standard imitation learning objectives may still produce asymmetric or inconsistent behaviors under symmetric observations, which can degrade coordination quality and robustness~\cite{drolet2024}.

Recent work has begun to explore symmetry and equivariance in robot learning, for example through symmetry-aware architectures~\cite{li2025morphologically, apraez2025morphological}, equivariant representations~\cite{zhang2025canonical, equidiff, equiform}, or carefully designed data augmentation schemes~\cite{bharadhwaj2023roboagent, chen2024roviaug}.
While these approaches have demonstrated promising results, they are often tightly coupled to specific model designs, observation modalities, or action parameterizations, such as particular network architectures or state representations.
As a result, it remains unclear how symmetry can be incorporated as a general and reusable inductive bias that can be applied consistently across diverse imitation learning pipelines.

In this paper, we propose EquiBim, a symmetry-aware training framework for dual-arm robot learning that explicitly enforces bilateral equivariance between observations and actions.
We formulate symmetry as a group action on observation and action spaces, and introduce an equivariance-based regularization that constrains policy predictions to transform consistently under symmetric transformations.
Crucially, EquiBim is model-agnostic and can be seamlessly integrated into existing imitation learning pipelines without modifying network architectures.
We evaluate the proposed framework on RoboTwin~\cite{chen2025robotwin}, a dual-arm robotic platform with symmetric kinematics, and demonstrate its effectiveness across diverse combinations of observation modalities, including point clouds and images, as well as action representations in both end-effector and joint spaces.
Experimental results in simulation and on real hardware show that explicitly enforcing symmetry provides a simple yet effective inductive bias for robust dual-arm robot learning.

In summary, this work makes three main contributions:
\begin{enumerate}
\item We propose EquiBim, a plug-and-play symmetry-equivariant regularization framework for bimanual imitation learning.
\item We formulate bilateral symmetry consistently across observations and actions, enabling applicability to diverse observation modalities and action spaces.
\item Extensive simulation and real-world experiments demonstrate improved robustness and generalization.
\end{enumerate}

\section{RELATED WORK}
\subsection{Bimanual Manipulation Learning}
Bimanual manipulation has long been recognized as a fundamental challenge in robotics, as it requires precise coordination between two manipulators while reasoning about shared object dynamics, contact interactions, and coupled kinematic constraints~\cite{smith2012dual, shaw2024bimanual}.
Compared to single-arm manipulation, dual-arm tasks introduce additional complexity in action coordination and significantly expand the space of feasible solutions.

Early approaches to bimanual manipulation often relied on task-specific controllers, manually designed coordination heuristics, or tightly engineered motion primitives~\cite{smith2012dual}.
While effective in structured settings, these methods typically lack scalability and struggle to generalize to new tasks or object configurations.
More recently, learning-based approaches have shown promising results in bimanual manipulation.
Imitation learning and reinforcement learning methods have been used to acquire coordinated dual-arm behaviors directly from demonstrations or interaction data~\cite{heng2025vitacformer, jiang2025dexmimicgen, xie2020deep, tan2025anypos, liu2021collaborative}. Such advances have enabled learning-based systems to handle a wider range of objects, scenes, and manipulation skills.

Despite this progress, most existing bimanual learning methods treat symmetry only implicitly.
Symmetric behaviors are typically learned through the data distribution rather than being explicitly enforced by the learning objective.
As a result, policies may exhibit asymmetric or inconsistent responses when presented with symmetric task configurations, especially in settings with limited demonstrations or imbalanced data coverage.
This limitation motivates the need for learning frameworks that can explicitly encode and enforce symmetry in dual-arm manipulation.

\subsection{Symmetry and Structural Priors in Robot Learning}
Incorporating symmetry and structural priors has been an active area of research in robot learning and representation learning~\cite{li2025morphologically, apraez2025morphological}.
Prior work has explored various mechanisms for leveraging known geometric or physical structures, including group-equivariant neural network architectures~\cite{equidiff, Equibot}, data canonicalization and augmentation strategies~\cite{zhang2025canonical, equiform, chen2024roviaug}, invariant feature representations~\cite{actionflow, ipa}.
By exploiting symmetries such as rotations, reflections, or permutations, these approaches have demonstrated improvements in generalization and sample efficiency.

Group-equivariant models, in particular, have been extensively studied in perception and representation learning, where equivariance to transformations can be encoded directly into network architectures.
In robotics, related ideas have been applied to problems such as state estimation~\cite{zhu2025residual}, motion planning~\cite{zhao2024mathrm}, and manipulation~\cite{li2025morphologically}, often through carefully designed network modules or handcrafted transformation rules tailored to specific tasks or input modalities.

While effective in their respective domains, many symmetry-aware approaches are typically instantiated through modality-specific representations or architecture-dependent designs~\cite{li2025morphologically, apraez2025morphological}.
This design choice can introduce practical challenges when transferring symmetry mechanisms across different robot learning pipelines, especially across varying observation spaces and action parameterizations.

In contrast, our work is motivated by the inherent bilateral symmetry in dual-arm manipulation, where symmetric observations and actions naturally arise from robot morphology and task structure.
Rather than encoding symmetry through specialized architectures, we formulate symmetry as a prediction-level consistency constraint during training, explicitly enforcing equivariant behavior under left-right symmetric transformations.
Unlike conventional mirror-based data augmentation, EquiBim enforces symmetry through prediction-level consistency and does not require mirrored supervision during training.
By decoupling symmetry enforcement from model design, the proposed EquiBim can be seamlessly integrated as a plug-and-play training module into existing imitation learning pipelines, making it particularly well suited for dual-arm robot learning across diverse observation modalities and action representations.

\section{SYMMETRY-AWARE LEARNING FRAMEWORK}
\subsection{Problem Setup and Bimanual Symmetry}
We study the problem of learning a robotic control policy $\pi$ via behavior cloning.
Given a history of observations
$\mathcal{O} = \{O_{t-m+1}, \ldots, O_t\}$ over the past $m$ time steps,
the goal is to predict a sequence of future actions
$\mathcal{A} = \{\mathbf{a}_t, \ldots, \mathbf{a}_{t+n-1}\}$ over a horizon of $n$ steps:
\begin{equation*}
\label{eq:map}
\pi: \mathcal{O} \rightarrow \mathcal{A},
\end{equation*}
where $t$ denotes the current time step, such that the learned policy imitates expert demonstrations.
Each observation $O_t$ may consist of heterogeneous sensory inputs,
including visual observations (e.g., images or point clouds)
and robot state information (e.g., end-effector states or joint configurations).
We make no assumptions about the specific observation modality or action representation,
and treat the policy as a generic mapping from observations to control commands.

We focus on bimanual manipulation tasks, where a robot operates two arms
to jointly perform a manipulation objective.
In many such tasks, the physical setup and task specification exhibit
inherent symmetry between the left and right arms.
Intuitively, exchanging the roles of the two arms while applying a corresponding
geometric reflection to the scene results in an equivalent task instance
and an equally valid expert behavior. 
We denote this structural property using a symmetry transformation
$\mathcal{S}(\cdot)$, which acts consistently on both observations and actions.
Applying $\mathcal{S}$ to an observation yields a symmetrically transformed observation,
while applying $\mathcal{S}$ to an action produces the corresponding transformed control.
For bimanual tasks with left--right symmetry, expert demonstrations are equivariant
under $\mathcal{S}$, reflecting a task-level prior that is independent of
the policy architecture or input representation.

\subsection{Symmetry-Consistent Regularization}
\begin{figure}[t]
    \centering
    \includegraphics[width=1.0\linewidth]{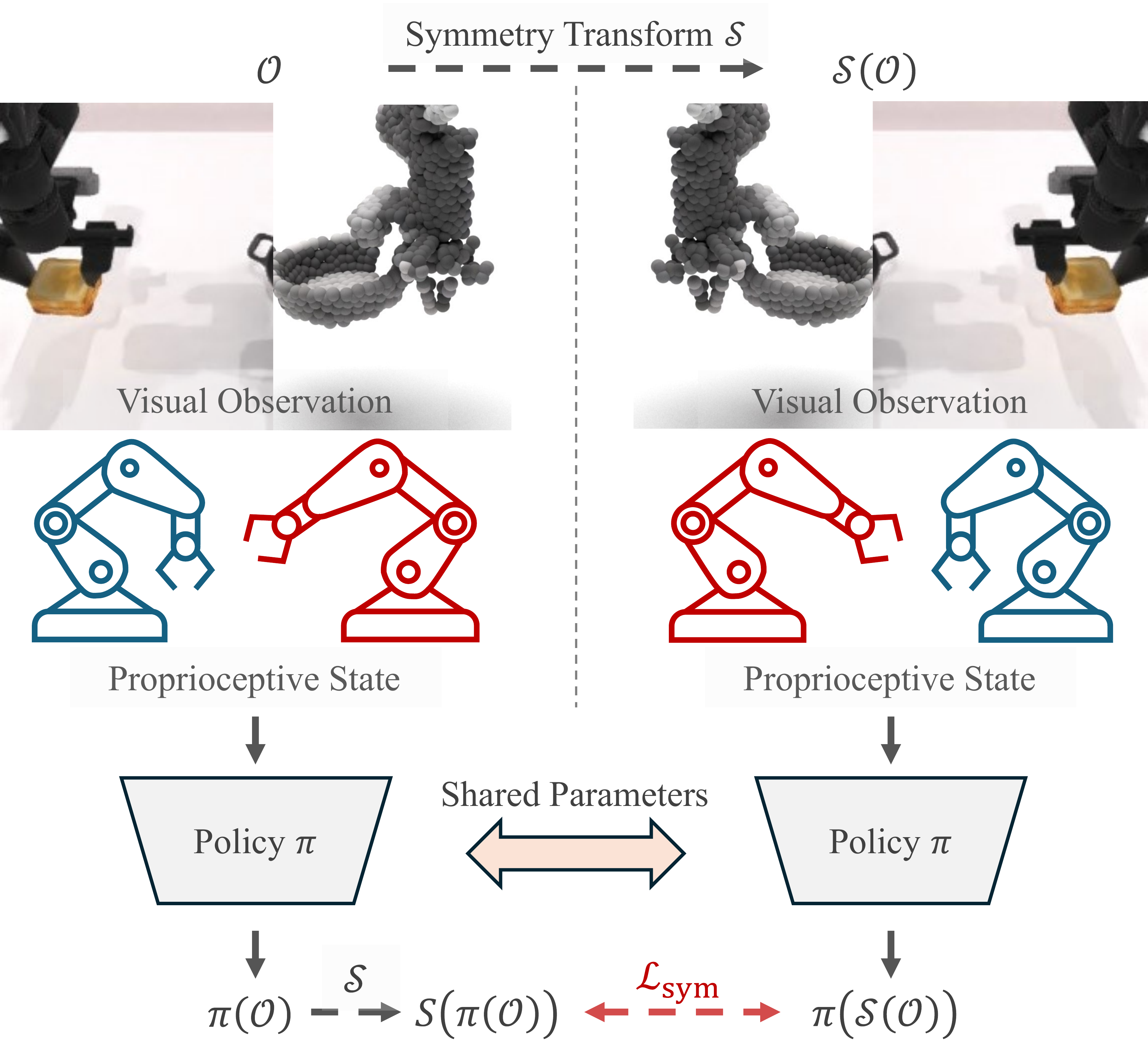}
    \caption{Overview of the proposed EquiBim framework. Given an observation $\mathcal{O}$ and its symmetrically transformed counterpart $\mathcal{S}(\mathcal{O})$, where $\mathcal{S}$ denotes a left-right reflection defined in the image frame, both inputs are processed by a shared policy $\pi$. The policy takes visual observations together with proprioceptive states as input and produces action predictions $\pi(\mathcal{O})$ and $\pi(\mathcal{S}(\mathcal{O}))$. A symmetry-equivariant loss $\mathcal{L}_{\mathrm{sym}}$ enforces the predicted actions to transform consistently under $\mathcal{S}$ during training.}
    \label{fig:method}
\end{figure}
An overview of the proposed EquiBim training pipeline is shown in Fig.~\ref{fig:method}.
Motivated by the inherent symmetry present in bimanual manipulation tasks, EquiBim introduces a symmetry-consistent regularization that encourages equivariant policy behavior under the symmetry transformation $\mathcal{S}$.
Rather than enforcing symmetry through architectural constraints,
we impose a consistency objective on the policy outputs, enabling plug-and-play
integration with existing imitation learning frameworks.

Given an observation history $\mathcal{O}$ and its symmetrically transformed
counterpart $\mathcal{S}(\mathcal{O})$, a symmetry-aware policy should produce
predictions that are consistent under $\mathcal{S}$.
Specifically, the action sequence predicted from the transformed observation,
$\pi(\mathcal{S}(\mathcal{O}))$, should match the symmetrically transformed
prediction from the original observation, $\mathcal{S}(\pi(\mathcal{O}))$.
This equivariant consistency can be enforced through the following regularization:
\begin{equation}
\mathcal{L}_{\text{sym}} =
\left\| \pi(\mathcal{S}(\mathcal{O})) - \mathcal{S}(\pi(\mathcal{O})) \right\|_2.
\end{equation}
Intuitively, this objective encourages the policy to produce compatible actions
for symmetrically equivalent task instances, even though the policy parameters
and inference process remain unchanged.
The regularization operates purely at the output level and does not require
additional supervision or explicit symmetry annotations.

Importantly, EquiBim’s symmetry-consistent regularization is agnostic to the underlying policy architecture, observation modality, and action representation.
As long as the symmetry transformation $\mathcal{S}$ can be consistently defined for the chosen inputs and outputs, the regularization can be applied without modifying the policy structure or training pipeline.
EquiBim introduces negligible computational overhead during training, requiring only one additional forward pass to evaluate the symmetry-consistent loss, while incurring no additional computation during inference.
This flexibility makes EquiBim particularly suitable for bimanual manipulation settings, where symmetry is a natural yet often underexploited task prior.

\subsection{Symmetry Transformation across Modalities}
\label{sec:symmetry}
\begin{figure*}[t]
    \centering
    \includegraphics[width=1.0\linewidth]{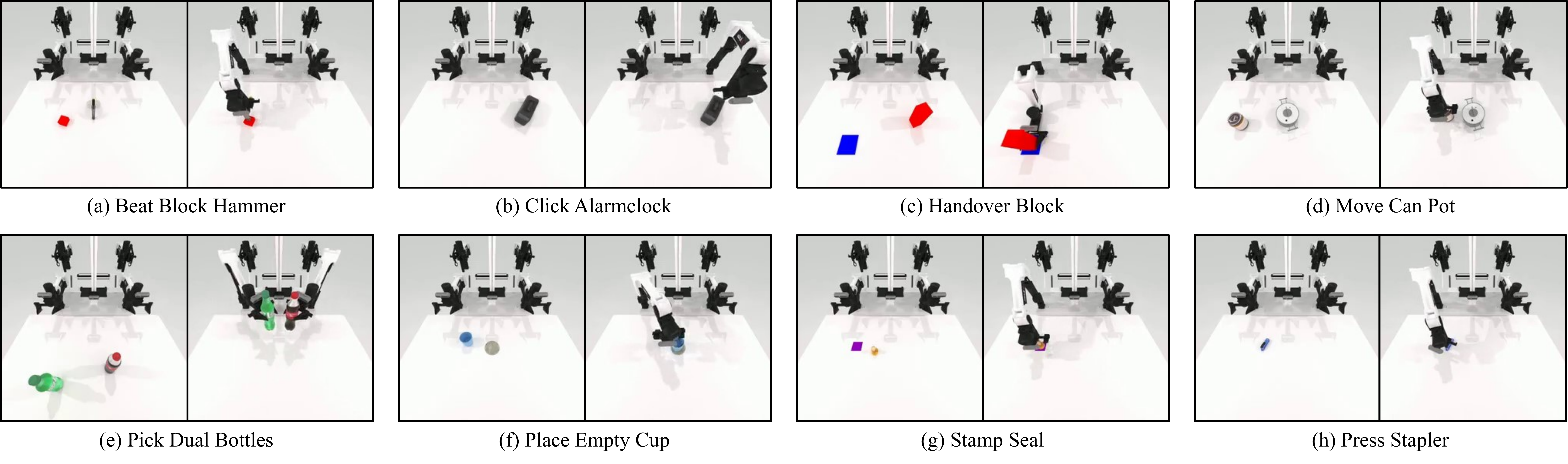}
    \caption{Visualization of eight simulated bimanual manipulation tasks used for evaluation. Subfigures (a)-(g) are adapted from RoboTwin~\cite{chen2025robotwin}. For each task, the left image shows the initial state and the right image shows the goal state.}
    \label{fig:sim}
\end{figure*}

The proposed symmetry-consistent regularization is broadly applicable across different observation modalities and action representations commonly used in bimanual manipulation.
In this work, we consider visual observations in the form of images or point clouds, and action spaces defined either in end-effector pose space or joint space.
All combinations of these observation and action choices are accommodated under a unified symmetry transformation $\mathcal{S}$.

We define the symmetry transformation $\mathcal{S}$ with respect to the left--right (lateral) direction of the robot workspace, corresponding to a
sagittal reflection commonly assumed in bimanual manipulation.
In typical bimanual robotic setups, visual observations are provided by a head camera mounted near the center of the robot and facing the shared workspace of the two arms.
Under this configuration, the horizontal axis of the image aligns with the left--right direction of the robot workspace, yielding a visually symmetric view under arm exchange.
Throughout this work, we therefore assume that the image horizontal axis corresponds to the robot's lateral ($y$) direction, and define all symmetry operations on observations and actions consistently under this convention.

For image-based observations, $\mathcal{S}$ corresponds to a horizontal flip of the image.
For point cloud observations, points are first transformed from the robot (or world) frame into the image coordinate frame using known camera extrinsics.
The symmetry transformation is then applied as a reflection along the lateral axis, followed by an inverse transformation back to the original frame. Formally, for a point $\mathbf{p}$,
\begin{equation}
\mathcal{S}(\mathbf{p}) =
\mathbf{T}_{\mathrm{cam}}^{-1}
\mathbf{R}_{y}
\mathbf{T}_{\mathrm{cam}}
\mathbf{p},
\end{equation}
where $\mathbf{T}_{\mathrm{cam}}$ denotes the transformation from the robot frame to the image frame, and $\mathbf{R}_{y} = \mathrm{diag}(1, -1, 1)$ represents reflection across the robot's sagittal plane.

A similar procedure is applied to end-effector actions. End-effector poses are first expressed in the image (or robot base) coordinate frame, where the same sagittal reflection is applied to both position and orientation, and are then transformed back to the original control frame.
This ensures that symmetry consistency is enforced in a geometrically coherent manner across perception and control.

For joint-space actions, the symmetry transformation $\mathcal{S}$ is defined based on the kinematic structure of the robot.
Each joint is assigned a symmetry sign according to whether its motion changes the left--right spatial configuration under arm exchange, as determined from the robot URDF model.
The transformed joint command is obtained by applying joint-wise sign flips:
\begin{equation}
\mathcal{S}(\mathbf{q}) = \mathbf{D}\mathbf{q},
\end{equation}
where $\mathbf{q}$ denotes the joint action vector and $\mathbf{D}$ is a diagonal matrix encoding joint-specific symmetry signs.

Throughout this work, actions are parameterized in absolute form (e.g., absolute end-effector poses or joint configurations), ensuring that the symmetry transformation $\mathcal{S}$ is well-defined and directly comparable across different action spaces.
By defining $\mathcal{S}$ consistently across observation and action representations, the same symmetry-consistent regularization can be applied without modification to policies operating on image or point cloud inputs and producing actions in either end-effector or joint space.
This unified treatment enables symmetry-aware learning to be seamlessly integrated into a wide range of bimanual manipulation pipelines.

\section{EXPERIMENTS}
\subsection{Experimental Setup}
We evaluate EquiBim under a controlled and modular experimental setup to isolate the effect of symmetry enforcement in bimanual manipulation.
All experiments are conducted on the RoboTwin~\cite{chen2025robotwin} benchmark, a dual-arm robotic simulation suite with symmetric kinematics and task structure, which makes it particularly suitable for studying bilateral symmetry.
As illustrated in Fig.~\ref{fig:sim}, we select eight representative tasks that exhibit natural left--right symmetry in both scene configuration and manipulation strategy.
In these tasks, exchanging the roles of the two arms leads to equivalent task instances and solution behaviors.

To assess the generality of the proposed symmetry regularization, we consider multiple combinations of observation modalities and action representations.
Specifically, we evaluate policies using either image-based or point cloud--based observations, paired with action spaces defined in either end-effector pose space or joint space.
This results in four observation--action configurations, enabling us to examine whether symmetry consistency can be enforced uniformly across different perception and control interfaces.

As baseline policies, we adopt the image-based Diffusion Policy~\cite{diffusion_policy} and the point cloud-based DP3~\cite{DP3}, which are widely used visual imitation learning approaches for bimanual manipulation.
For each backbone and observation-action configuration, we compare the original policy against its EquiBim-augmented counterpart.
Importantly, EquiBim is applied without modifying the underlying policy architecture, optimization procedure, or training pipeline, ensuring that any performance differences can be attributed solely to the imposed symmetry-equivariant regularization.
Beyond simulation, we further validate the approach on a real-world dual-arm robotic system, with details provided in Section~\ref{sec:realworld}.

\subsection{Effect of Symmetry Regularization in Simulation}
\begin{table}[t]
\centering
\footnotesize
\caption{Average success rate (\%) across observation--action settings on RoboTwin.}
\label{tab:overall_EquiBim}
\begin{tabular*}{\columnwidth}{@{\extracolsep{\fill}}l l c c c}
\toprule
\textbf{} & \textbf{Observation + Action} & \textbf{Baseline} & \textbf{+ EquiBim} & $\boldsymbol{\Delta}$ \\
\midrule
\multirow{2}{*}{DP} 
    & Image + Joint & 34.1 & \textbf{43.6} & \textbf{+9.5} \\
    & Image + EE    & 37.3 & \textbf{40.0} & \textbf{+2.7} \\
\midrule
\multirow{2}{*}{DP3} 
    & Point Cloud + Joint & 73.5 & \textbf{77.9} & \textbf{+4.4} \\
    & Point Cloud + EE    & 74.5 & \textbf{77.8} & \textbf{+3.3} \\
\bottomrule
\end{tabular*}
\end{table}

\begin{table*}[t]
\centering
\footnotesize
\renewcommand{\arraystretch}{1.15}
\setlength{\tabcolsep}{4pt}
\caption{Per-task performance gains from symmetry regularization ($\Delta$ success rate, \%).}
\label{tab:task_gain_EquiBim}
\begin{tabular}{
l
p{3.0cm}
@{\hskip 10pt}c
@{\hskip 10pt}c
@{\hskip 10pt}c
@{\hskip 10pt}c
@{\hskip 10pt}c
@{\hskip 10pt}c
@{\hskip 10pt}c
@{\hskip 10pt}c
}
\toprule
\textbf{} &
\textbf{Observation + Action} &
\textbf{Beat} &
\textbf{Click} &
\textbf{Handover} &
\textbf{Move} &
\textbf{Pick} &
\textbf{Place} &
\textbf{Stamp} &
\textbf{Press} \\
\midrule

\multirow{2}{*}{DP}
& Image + Joint       
& +22 & $+$0 & +2  & +23 & $-$11 & +28 & +3  & +9  \\
& Image + EE          
& +44 & $-$6  & $-$15 & +9  & $-$13 & $-$2 & +1  & +4  \\

\midrule

\multirow{2}{*}{DP3}
& Point Cloud + Joint 
& +12 & +7  & $-$1 & $-$6 & +4  & +6  & $-$2 & +15 \\
& Point Cloud + EE    
& +4  & $+$5 & $+$1 & +3  & $-$5 & +9  & $-$1 & +12 \\

\bottomrule
\end{tabular}
\end{table*}
Based on the experimental setup described above, we evaluate the effect of EquiBim in simulation by comparing policies trained with and without the symmetry-consistent regularization under identical conditions.
Performance is measured by the average success rate across the eight RoboTwin tasks listed in Table~\ref{tab:overall_EquiBim}, where Joint denotes the joint-space action representation, while EE denotes the end-effector action representation.

As shown in Table~\ref{tab:overall_EquiBim}, EquiBim consistently improves performance across all observation--action settings. The largest gain is observed in the Image + Joint configuration, where the success rate increases from 34.1\% to 43.6\%, corresponding to an improvement of +9.5\%. Image + EE also improves by +2.7\%. In comparison, point cloud--based settings exhibit more moderate gains of +4.4\% (Point Cloud + Joint) and +3.3\% (Point Cloud + EE).

This trend can be attributed to the amount of spatial structure encoded in the representations. Point clouds explicitly capture 3D geometry, and end-effector actions are defined in Cartesian space, both of which provide strong geometric priors. In contrast, image observations lack explicit 3D structure, and joint-space actions do not directly reflect task-space spatial relationships. Consequently, policies trained with Image + Joint must implicitly infer spatial correspondences and bilateral consistency from weaker signals. In this setting, symmetry regularization introduces an additional structural prior that constrains the policy to behave consistently under symmetric transformations, effectively compensating for missing geometric information and leading to larger gains.

Table~\ref{tab:task_gain_EquiBim} further reveals per-task performance differences. EquiBim yields substantial improvements on tasks with clear bilateral structure or interchangeable arm roles, such as \textit{Beat Block Hammer}, \textit{Move Can Pot}, \textit{Place Empty Cup}, and \textit{Press Stapler}. These tasks involve symmetric spatial layouts or coordinated bimanual interactions, where enforcing symmetry consistency reduces policy variance and improves generalization.

We also observe performance drops on several tasks, notably \textit{Handover Block} and \textit{Pick Dual Bottles}. Although these tasks appear spatially symmetric, their optimal control strategies may exhibit instance-level asymmetry. For \textit{Handover Block}, the transfer behavior is temporally structured: one arm typically stabilizes the object while the other performs the primary manipulation, creating a functional role asymmetry despite geometric symmetry. Similarly, \textit{Pick Dual Bottles} may involve slight differences in object pose, contact conditions, or grasp ordering that require asymmetric force application or timing. In these cases, symmetry regularization can partially suppress useful asymmetric adaptations, leading to moderate performance degradation.

Importantly, this does not imply that handover-type tasks are inherently incompatible with symmetry regularization. In our real-world banana handover experiment, EquiBim improves performance over the baseline. This suggests that when task-level symmetry is dominant, symmetry consistency helps stabilize coordination and improve robustness. The simulation failures likely arise from task-specific interaction details or stochastic variations that introduce small but important asymmetries. Overall, these results indicate that symmetry regularization is most beneficial when task symmetry aligns with the dominant control structure, while remaining generally robust across diverse manipulation scenarios.

\subsection{Real-World Evaluation}
\label{sec:realworld}
\begin{figure}[t]
    \centering
    \includegraphics[width=1.0\linewidth]{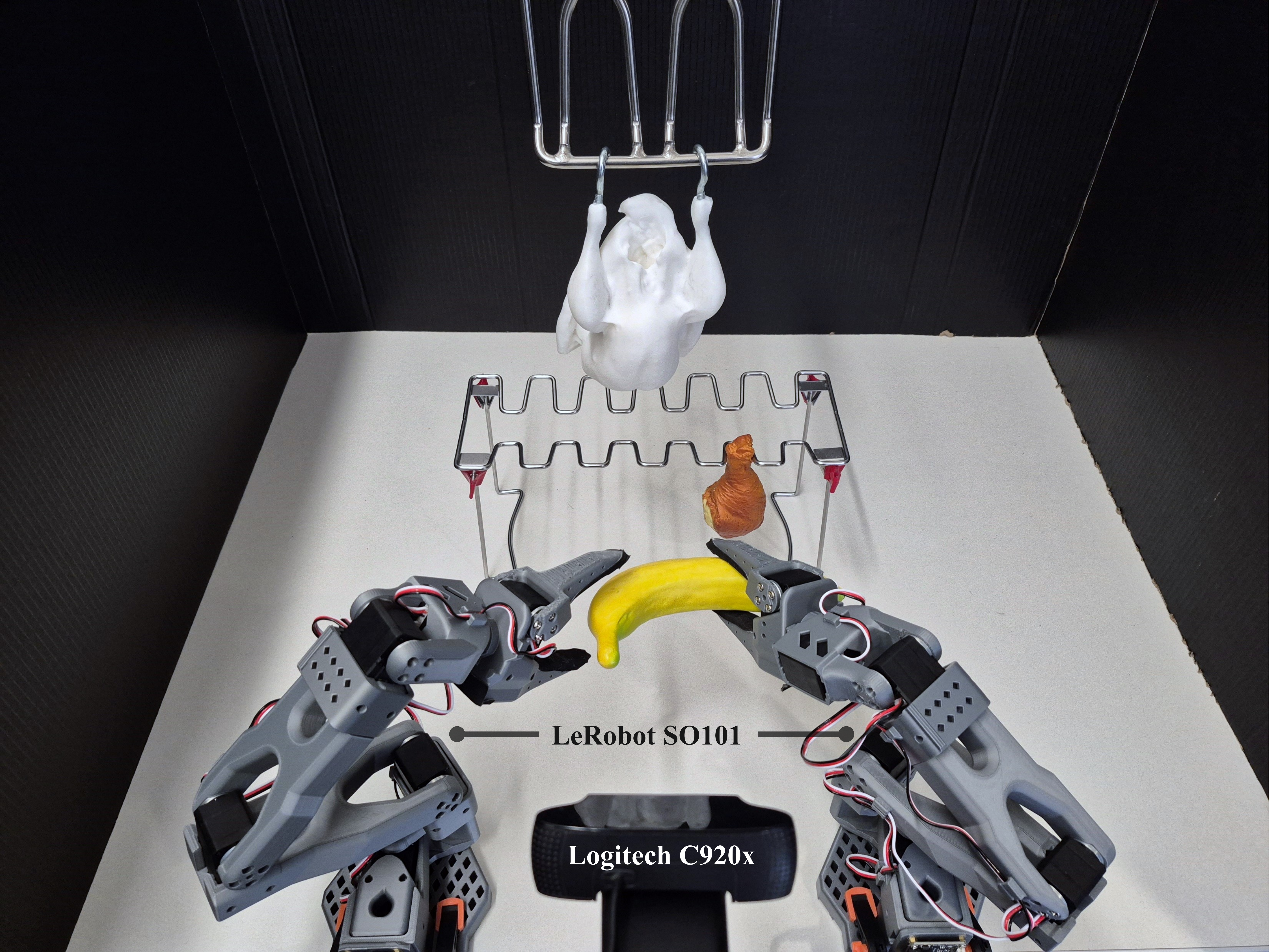}
    \caption{Overview of the real-world bimanual manipulation setup. Two LeRobot SO101 robotic arms operate in a shared workspace and are observed by a Logitech C920x camera. The system is evaluated on object handover and hook-hanging tasks under a top-down camera configuration.}
    \label{fig:realworld}
\end{figure}

\begin{figure*}[t]
    \centering
    \includegraphics[width=1.0\linewidth]{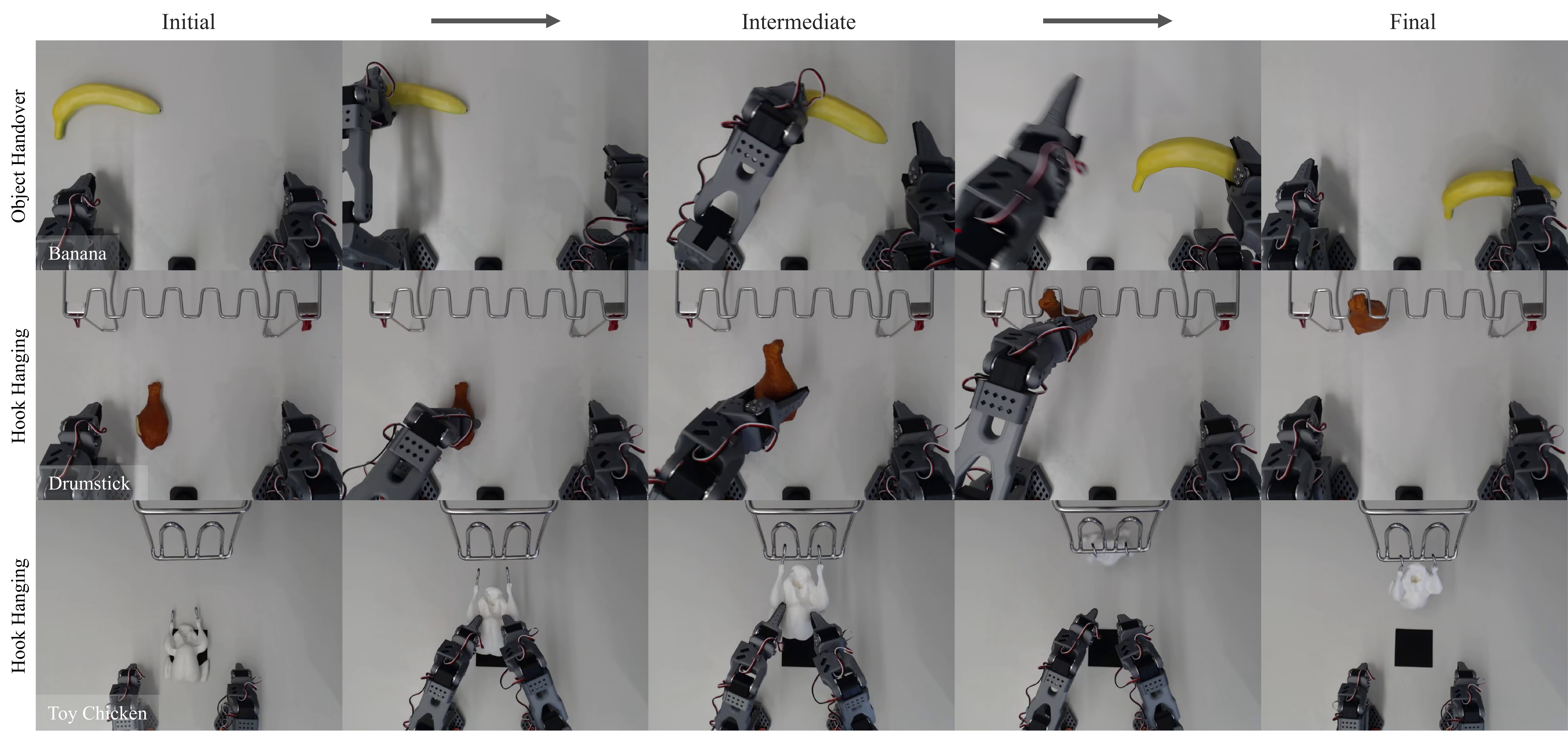}
    \caption{Real-world task executions on the bimanual LeRobot platform. The figure presents three tasks, namely Object Handover (Banana), Hook Hanging (Drumstick), and Hook Hanging (Toy Chicken), each depicted across three stages: initial, intermediate, and final. Each row shows the temporal progression of the manipulation process under a top-down RGB camera view.}
    \label{fig:timeline}
\end{figure*}

\begin{figure}[t]
    \centering
    \includegraphics[width=1.0\linewidth]{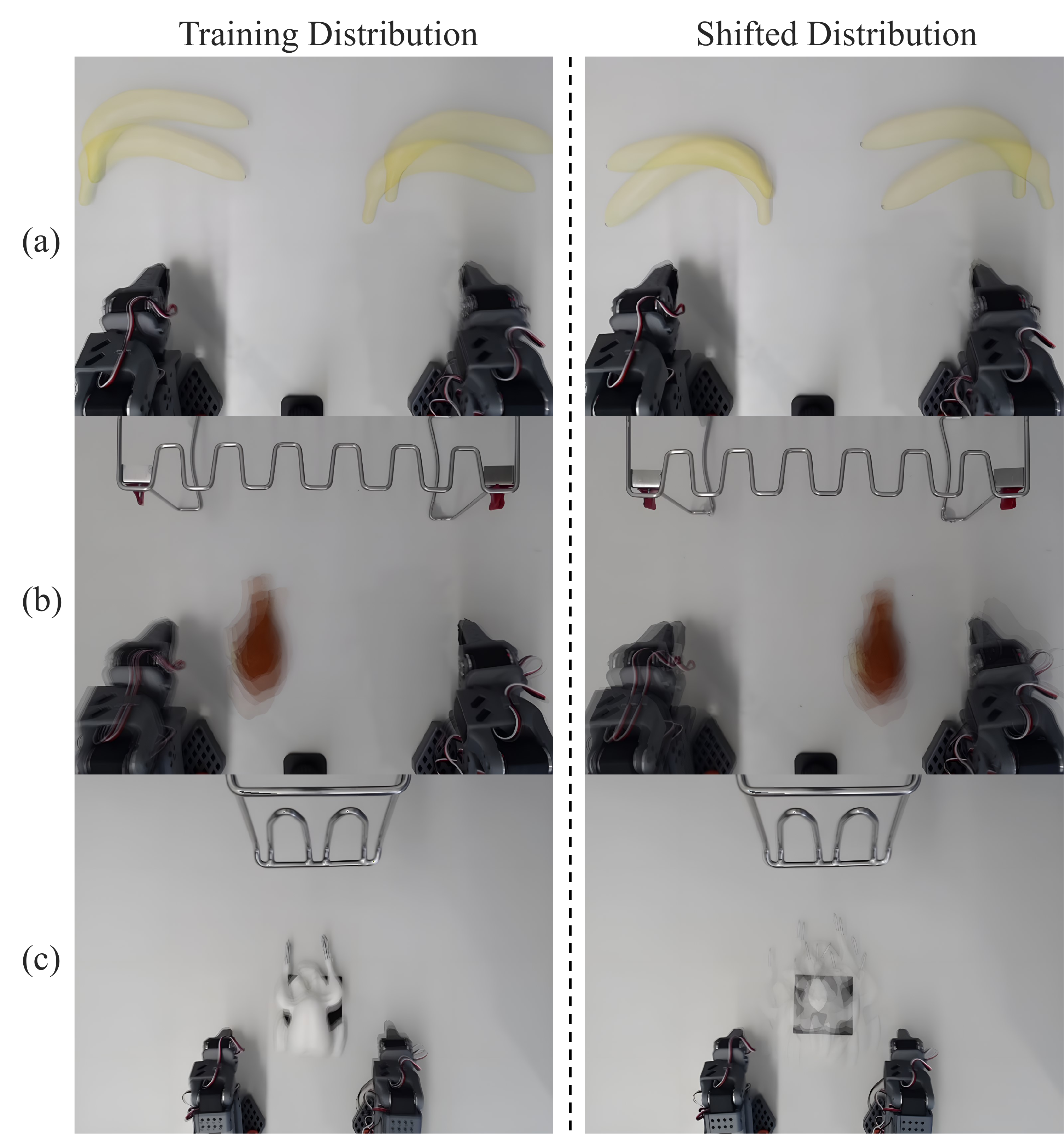}
    \caption{Generalization under distribution shift on the real-world bimanual platform. Rollouts are shown under the training distribution (left) and shifted conditions (right) for object handover, drumstick hook hanging, and toy chicken hook hanging. Each visualization overlays multiple rollouts (out of 10 total trials per setting) using transparency to illustrate trajectory variability. For clarity, only a subset of representative rollouts is displayed. The shifts include symmetric orientation changes, demonstration quality variation, and randomized initialization.}
    \label{fig:overlap}
\end{figure}
\label{sec:overlap}

\begin{table*}[t]
\centering
\footnotesize
\renewcommand{\arraystretch}{1.15}
\setlength{\tabcolsep}{4pt}
\caption{Success rate in real-world experiments under training and shifted object distributions.}
\label{tab:real_world_results}

\begin{tabular}{
p{2.2cm}
*{6}{>{\centering\arraybackslash}p{1.8cm}}
}
\toprule
\textbf{Method} &
\multicolumn{3}{c}{\textbf{Training Distribution}} &
\multicolumn{3}{c}{\textbf{Shifted Distribution}} \\

\cmidrule(lr){2-4}
\cmidrule(lr){5-7}

& \textbf{Banana} 
& \textbf{Drumstick} 
& \textbf{Toy Chicken}
& \textbf{Banana} 
& \textbf{Drumstick} 
& \textbf{Toy Chicken} \\

\midrule

ACT 
& 3/10 & 3/10 & 8/10 
& 0/10 & 1/10 & 4/10 \\

ACT + EquiBim 
& \textbf{6/10} & \textbf{3/10} & \textbf{9/10}
& \textbf{5/10} & \textbf{4/10} & \textbf{6/10} \\

\bottomrule
\end{tabular}

\vspace{0.3em}
{\footnotesize All real-world policies were evaluated over 10 independent trials per object using image observations and joint-space actions.}
\end{table*}

We validate EquiBim on a real-world dual-arm LeRobot SO101~\cite{cadene2026lerobot} platform, as shown in Fig.~\ref{fig:realworld}. The two arms are symmetrically mounted and operate within a shared workspace. A Logitech C920x RGB camera is positioned near the center of the platform, mimicking a head-mounted configuration. Under this setup, the horizontal image axis aligns with the left--right direction of the workspace, producing visually symmetric observations under arm exchange. This design matches the symmetry assumptions described in the method section and enables controlled evaluation of symmetry consistency in real manipulation. Human demonstrations are collected using a leader-follower teleoperation system, where two leader arms independently control the corresponding follower arms through direct joint-space mapping.

We evaluate three tasks. (1) \textbf{Banana Handover}: depending on whether the banana is placed on the left or right side, the corresponding arm grasps it and transfers it to the opposite arm. (2) \textbf{Drumstick Hook Hanging}: the arm on the same side as the drumstick grasps it and hangs it onto the rack. (3) \textbf{Toy Chicken Hook Hanging}: both arms cooperatively grasp the toy chicken and hang it onto the hook. The temporal progression of each task is illustrated in Fig.~\ref{fig:timeline}. For each task, we collect 50 human demonstrations. Policies are trained using ACT~\cite{zhao2023learning}, taking as input the RGB image and the joint angles of both arms.

Under the training distribution (Fig.~\ref{fig:overlap}, left; Table~\ref{tab:real_world_results}), EquiBim improves performance on tasks exhibiting clear symmetric structure. The overlap visualizations aggregate multiple rollouts (10 trials per setting) using semi-transparency to illustrate trajectory variability; for clarity, only a subset of representative rollouts is displayed. For Banana Handover, ACT achieves 3/10 success, whereas ACT + EquiBim reaches 6/10, corresponding to a 30\% absolute improvement. For Drumstick and Toy Chicken hook hanging, where rollout configurations closely resemble the demonstrations, both methods achieve comparable performance, with EquiBim providing consistent but smaller gains.

To further evaluate generalization, we introduce distribution shifts (Fig.~\ref{fig:overlap}, right; Table~\ref{tab:real_world_results}). As shown in Fig.~\ref{fig:overlap}(a), for Banana Handover, the banana orientation is mirrored relative to training and its placement is varied across sides. Under this symmetric shift, the baseline ACT fails (0/10), while ACT + EquiBim achieves 5/10 success. This demonstrates that symmetry regularization enables the policy to transfer learned left--right interaction patterns across mirrored configurations, significantly improving robustness.

As shown in Fig.~\ref{fig:overlap}(b), in Drumstick Hook Hanging, demonstration quality is intentionally asymmetric: demonstrations are nearly perfect when the drumstick is placed on the left, whereas 7 out of 25 demonstrations fail when placed on the right. Under this uneven supervision, ACT + EquiBim achieves 4/10 success compared to 1/10 for ACT under distribution shift. This suggests that symmetry regularization leverages the stronger left-side demonstrations to regularize and compensate for weaker right-side data, effectively transferring symmetric structure across arms.

Finally, as shown in Fig.~\ref{fig:overlap}(c), for Toy Chicken Hook Hanging, we introduce broader randomized initializations. Under this positional shift, ACT + EquiBim achieves 6/10 success compared to 4/10 for ACT, corresponding to a 20\% improvement. This indicates improved generalization under varying initial conditions.

Overall, the real-world results demonstrate that symmetry-consistent regularization enhances robustness and generalization in visually symmetric bimanual manipulation settings, yielding consistent improvements across both training and shifted distributions.

\section{CONCLUSIONS}
In this paper, we presented EquiBim, a symmetry-consistent regularization framework for bimanual imitation learning that explicitly enforces bilateral equivariance between observations and actions.
By formulating symmetry as a group action jointly acting on observation and action spaces, EquiBim provides a simple, plug-and-play inductive bias that can be seamlessly integrated into existing imitation learning pipelines without architectural modifications.

Extensive simulation experiments on the RoboTwin benchmark demonstrate that symmetry-consistent regularization consistently improves performance across diverse observation modalities and action representations.
Real-world evaluations on a dual-arm robotic platform further confirm enhanced coordination and robustness under both nominal and shifted conditions.

Future work will investigate extending EquiBim to more general camera configurations by relaxing the current camera alignment assumption and supporting arbitrary viewpoints. Another promising direction is to extend the framework to more complex robotic embodiments, such as dexterous hands, by defining appropriate symmetry transformations for higher-dimensional action spaces.
\addtolength{\textheight}{-10cm}   




\bibliographystyle{unsrt}
\bibliography{references}

\end{document}